\documentclass[conference]{IEEEtran}
\IEEEoverridecommandlockouts
\usepackage{cite}
\usepackage{amsmath,amssymb,amsfonts}
\usepackage{algorithmic}
\usepackage{graphicx}
\usepackage{adjustbox}
\usepackage{multirow}
\usepackage{textcomp}
\usepackage{color,soul}
\usepackage{xcolor}
\def\BibTeX{{\rm B\kern-.05em{\sc i\kern-.025em b}\kern-.08em
    T\kern-.1667em\lower.7ex\hbox{E}\kern-.125emX}}
\begin{document}

\title{ECG-Adv-GAN: Detecting ECG Adversarial Examples with Conditional Generative Adversarial Networks}

\iffalse
\title{Conference Paper Title*\\
{\footnotesize \textsuperscript{*}Note: Sub-titles are not captured in Xplore and
should not be used}
\thanks{Identify applicable funding agency here. If none, delete this.}
}
\fi

\author{\IEEEauthorblockN {Khondker Fariha Hossain\IEEEauthorrefmark{1}, Sharif Amit Kamran\IEEEauthorrefmark{2}, Alireza Tavakkoli\IEEEauthorrefmark{3}, Lei Pan\IEEEauthorrefmark{4}, Xingjun Ma\IEEEauthorrefmark{5},\\ Sutharshan Rajasegarar\IEEEauthorrefmark{6}, Chandan Karmaker\IEEEauthorrefmark{7}}

\IEEEauthorblockA{\IEEEauthorrefmark{1}\IEEEauthorrefmark{2}\IEEEauthorrefmark{3}
\textit{University of Nevada, Reno},
NV, USA  \\
\IEEEauthorrefmark{4}\IEEEauthorrefmark{5}\IEEEauthorrefmark{6}\IEEEauthorrefmark{7}
\textit{Deakin University},
Australia \\
khondkerfarihah@nevada.unr.edu\IEEEauthorrefmark{1}, skamran@nevada.unr.edu\IEEEauthorrefmark{2}, tavakkol@unr.edu\IEEEauthorrefmark{3}, l.pan@deakin.edu.au\IEEEauthorrefmark{4},\\ daniel.ma@deakin.edu.au\IEEEauthorrefmark{5}, sutharshan.rajasegarar@deakin.edu.au\IEEEauthorrefmark{6}, karmakar@deakin.edu.au\IEEEauthorrefmark{7}}
}

\maketitle

\begin{abstract}
Electrocardiogram (ECG) acquisition requires an automated system and analysis pipeline for understanding specific rhythm irregularities. Deep neural networks have become a popular technique for tracing ECG signals, outperforming human experts. Despite this, convolutional neural networks are susceptible to adversarial examples that can misclassify ECG signals and decrease the model's precision. Moreover, they do not generalize well on the out-of-distribution dataset. The GAN architecture has been employed in recent works to synthesize adversarial ECG signals to increase existing training data. However, they use a disjointed CNN-based classification architecture to detect arrhythmia. Till now, no versatile architecture has been proposed that can detect adversarial examples and classify arrhythmia simultaneously. To alleviate this, we propose a novel Conditional Generative Adversarial Network to simultaneously generate ECG signals for different categories and detect cardiac abnormalities. Moreover, the model is conditioned on class-specific ECG signals to synthesize realistic adversarial examples. Consequently, we compare our architecture and show how it outperforms other classification models in normal/abnormal ECG signal detection by benchmarking real world and adversarial signals. 
\end{abstract}


\section{Introduction}
Electrocardiogram (ECG) is a re-polarization sequence in the human heart recorded over time using a standard non-invasive tool. It is used for clinically diagnosing cardiac diseases by understanding abnormalities in the signal space. In addition, automated arrhythmia recognition systems have become the standard as of late.  Many popular systems utilize Deep Neural networks (DNN) in embedded devices to analyze certain rhythm irregularities in patients. For example, the Medtronic Linq monitor uses an injectable chip, the iRhythm Zio monitor uses a wearable patch, whereas Apple Watch Series 4 monitors rhythm irregularities from the wrist \cite{rajpurkar2017cardiologist}. Even insurance companies and contractors partner with these device companies by providing subsidies and benefits to the customers. As there are many stakeholders involved financially, the reliability of these toolkits becomes questionable. Moreover, the companies have to develop creative ways to manage ECG readings as they become more susceptible to Adversarial Examples \cite{finlayson2019adversarial}.

Many recent architectures incorporate convolutional or recurrent LSTM-based blocks to find inherent manifold features of different arrhythmia beats \cite{mousavi2019inter,kachuee2018ecg,acharya2017deep}. However, despite recent advances in ECG classification, these methods do not generalize well on out-of-distribution and real-world data.  Another problem with deep-learning architecture is, they are susceptible to adversarial examples \cite{han2020deep}. Quite recently, there has been a couple of works that adopted GAN for generating synthetic ECG data \cite{golany2020improving,shaker2020generalization}. However, the first one \cite{golany2020improving} uses a disjointed architecture to classify arrhythmia, which cannot predict if the signal is real or fake.Additionally, the architecture only utilizes an unbounded adversarial loss, which results in a random signal generation overlooking any class-specific signal features. Also, the authors only evaluate their model on real ECG signals, providing no benchmarks for the adversarial ECG test data. For the second work, \cite{shaker2020generalization}, the authors focus more on the failures of ECG classification architectures in regards to adversarial attacks and provides no defense against such attacks.

\begin{figure*}[!t]
    \includegraphics[width=\textwidth]{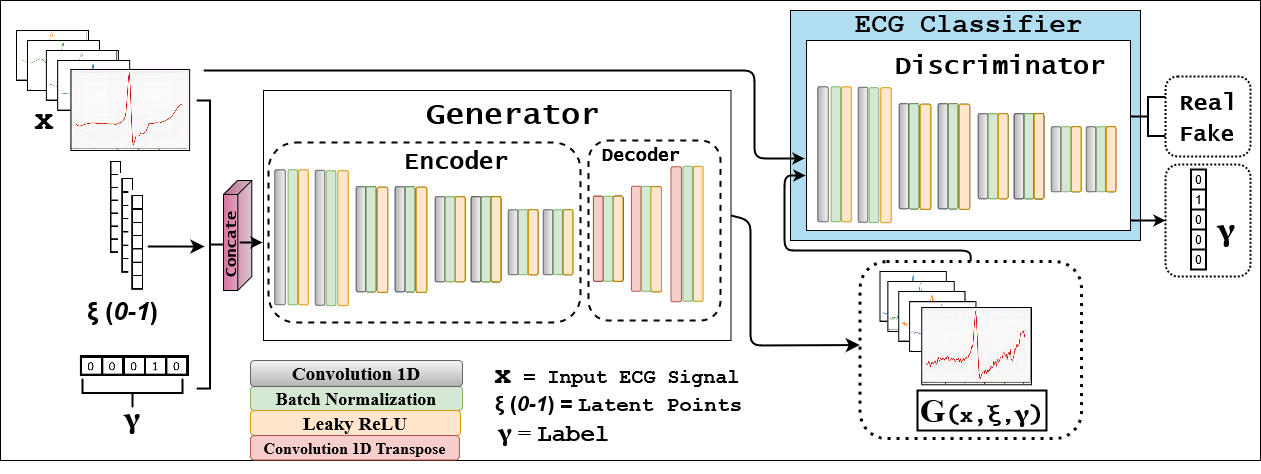}
    \caption{Proposed ECG-Adv-GAN consists of a single Generator and Discriminator where the Generator takes the Real ECG signals, a noise vector, and the class labels as input. The Generator consists of an encoder and decoder modules through which it outputs the synthesized adversarial signals. The Discriminator has only an encoding module that takes Real or Adversarial signal as input. It works as a classifier with two output modules, one for categorizing signals and the other for detecting adversarial examples. }
    \label{fig1}
\end{figure*}

To alleviate this, we propose a new conditional generative network, called ECG-Adv-GAN illustrated in Fig.~\ref{fig1}, that can simultaneously synthesize and classify both arrhythmia and adversarial examples as an end-to-end architecture. We benchmark our method with other state-of-the-art architecture on the Intra-patient, and Inter-patient datasets acquired from the MIT-BIH dataset \cite{moody2001impact}. Moreover, we also synthesize the adversarial test data using our generator and then evaluate our discriminator's performance comparing with other architectures. Consequently, by utilizing LS-GAN loss \cite{mao2017least}  and reconstruction loss, our generator synthesizes visually realistic ECG signals with small perturbations  that can be adopted in clinical applications and day-to-day activities.
\iffalse
This paper proposed a new multi-scale robust generative adversarial network that generates realistic adversarial ECG examples and detects real or fake ECG samples. Simultaneously, classify the cardiac abnormalities of the generated adversarial ECG sample with high precision.
\fi

\section{Related Works}

\subsection{Deep Learning Approaches}
Lately, deep learning architectures have become a go-to for achieving state-of-the-art accuracy for various ECG classification tasks \cite{ye2010arrhythmia,mousavi2019inter,acharya2017deep,kachuee2018ecg}. Earlier works incorporated Support vector machines (SVM) and Independent component analysis on top of morphological and dynamic features of ECG signals to group them into classes \cite{ye2010arrhythmia}. The problem with this approach is, it relies on handcrafted features extracted by human experts. In contrast, deep learning approaches classify different Heartbeats from ECG signals by automating the feature extraction process.  Convolutional neural network succeeds in extracting manifold features and generalizes better on large amounts of data. For example, Kachuee et al. \cite{kachuee2018ecg} employed a 1D CNN-based model to identify Cardiac abnormalities and obtained almost 94\% accuracy in detecting different arrhythmia diseases. Mousavi et al. \cite{mousavi2019inter} adopted a Seq-to-seq architecture consisting of bidirectional LSTM (Long short-time memory) to achieve the state-of-the-art result in arrhythmia classification from ECG signals. Although these architectures obtain high accuracy in classification tasks, they often fail to identify adversarial examples in the wild. One of our fundamental contributions is to illustrate the failure of such models to classify the correct arrhythmia category when exposed to realistic adversarial examples.

\subsection{Generative Adversarial Networks}

Generative Adversarial Networks (GAN) is a synthesis architecture for generating high-quality images, time-series data, and other modalities of imaging \cite{delaney2019synthesis,amit2021rv,kamran2021vtgan}. Generally, the architecture consists of a generator network and a discriminator network. Medical data are highly senstivie in nature, so it can be pretty challenging to synthesize artificial data from real data while retaining the inherence features and preserving privacy.  Many recent works incorporated GANs for ECG signal generation to tackle the problem of shortage of data \cite{delaney2019synthesis,murugesan2018ecgnet,shaker2020generalization}. Delaney et al. \cite{delaney2019synthesis} proposed a GAN architecture comprising a Bidirectional LSTM-as Generator and LSTM-CNN as Discriminator to synthesize ECG data. However, the authors only focused on synthesizing ECG signals using the generator and provided no evaluation of the discriminator network for the arrhythmia classification task. Shaker et al. \cite{shaker2020generalization} used a two-stage unsupervised approach for synthesizing and then classifying ECG signals. In the first stage, the authors utilized DC-GAN \cite{radford2015unsupervised} by taking Gaussian noise as input and generate random ECG signals for augmenting the training dataset. In the second stage, the authors utilized two separate networks for classifying the ECG signals into distinct categories.This approach has three underlying problems: 1) the generator model is not conditioned on unique class-specific signals and the authors utilized an unbounded adversarial loss, so the synthesis is random, 2) The classifier was not exposed to adversarial exmples while training, instead the author used a disjointed GAN training in first stage to augment the existing ECG data, 3) Utilizing two seperate architectures, 1) Generative Adversarial Network and, 2) ECG classifier is computationally expensive and the classifier can't determine if the ECG signals are real or adversarial. Moreover, the model's accuracy is relatively lower than other state-of-the-art ECG classification models. Our second contribution is, we propose a conditional GAN where the generator retains class-specific information by training it with a closely bounded reconstruction loss, and the discriminator network can simultaneously distinguish and classify real and adversarial ECG signals with high precision. 

\subsection{Clinical Significance}
Super realistic adversarial samples can be a temptation for companies in the context of drug and device approvals. In different clinical trials, regulatory bodies like FDA (Food and Drug Administration) affirmed interest in using algorithmic biomarkers as an endpoint \cite{finlayson2019adversarial}. In this context, adversarial examples can be a channel for companies to create favorable biased outcomes. For instance, a regulator requires a matched ECG signal for the patient before and after treatment. As a result, trialists could inject adversarial noise or create adversarial samples for post-treatment data, ensuring desired results \cite{finlayson2019adversarial,gottlieb2018fda}.  In such circumstances, an adversarial ECG detection system can play a pivotal role in identifying borderline fraudulent trials and, in the long run, can save lives. 

Moreover, the FDA recently released guidelines and tools to collect Real-World Data (RWD) from research participants \cite{klonoff2020new}. However, such extensive studies require a large amount of Patient-generated Health Data (PGHD), most of which are collected from wearable devices. It is both time-consuming and impractical for clinicians to analyze and do perturbation to address the data shortage \cite{hannun2019cardiologist}. So a more reliable method would be a generative architecture that can synthesize realistic ECG signals imitating distinct rhythmic irregularities. 

\section{Methodology}

In this paper, we proposed a conditional generative adversarial network (GAN) comprising a generator for producing realistic ECG signals and a discriminator for classifying the real and generated signal's output labels.
First, we discussed the proposed architecture in Section~\ref{parc}. In Section~\ref{WcA}, we discuss the objective function and loss weight distributions for each architecture related to proposed model.

\subsection{Proposed Architecture}
\label{parc}
Incorporating generative networks with an auxiliary classification module has been shown to produce high-quality image synthesis and highly accurate class prediction as observed in \cite{odena2017conditional}. Inspired by this, we adopt this feature in our architecture using a class conditioned generator and a multi-headed discriminator for categorical classification and adversarial example detection as visualized in Fig.~\ref{fig1}. The generator concatenates the original signal $x$, label $y$, and a noise vector $z$ as input and synthesizes $G(x,z,y)$ ECG signal of a distinct category. The noise vector, $z$ is smoothed with Gaussian filter with $\sigma=4$ before pushing as input.  The generators consist of multiple Convolution, Transposed Convolution, Batch-Norm, and Leaky-ReLU layers. We use convolution for downsampling $3\times$ and use transposed convolution to upsample $3\times$ again to make the dimension the same as the input. For normal convolution we use kernel size, $k=3$, stride, $s=1$ and padding, $p=1$. For downsampling convolution and transposed convolution we use kernel size, $k=3$, stride, $s=2$ and padding, $p=1$. Convolution and Transposed Convolutions are followed by Batch-normalization and Leaky-ReLU layers. The encoder consists of six convolution layers. The number of features are $[E1,E2,E3,E4,E5,E6]$= $[32,32,64,64,128,128]$. The decoder consists of three transposed convolution layers and they have $[D1,D2,D3]$ =$[128,64,32]$ number of features.  The generator has input and output dimension of $280\times1$ and incorporates Sigmoid activation as output. 

On the other hand, the discriminator takes both real signal $x$ and adversarial ECG signals $G(x,z,y)$ as input. It simultaneously predicts if the example is real or adversarial and classifies the signal as one of four categories. The discriminator consists of multiple Convolution, Batch-Norm, Leaky-ReLU, and Fully Connected layers. We use convolution for downsampling $4$ times. For convolution we use kernel size, $k=3$, stride, $s=1$ and padding, $p=1$, except for downsampling convolution where we use stride, $s=2$. The convolution layer is succeeded by Batch-normalization and Leaky-ReLU layers. After that, we use two fully connected layers.  The encoder consists of eight convolution layers and two dense layers. The number of features are for each layer is $[E1,E2,E3,E4,E5,E6,E7,E8,E9,E10]$= $[16,16,32,32,128,128,256,64,32]$. We use two output activation: classification with Softmax for four classes and Sigmoid for real/adversarial prediction.

\subsection{Weighted Cost Function and Adversarial Loss}
\label{WcA}

We use LSGAN \cite{mao2017least} for calculating the adversarial loss and training our Generative Network. The objective function for our conditional GAN is given in Eq.~\ref{eq1}. 
\begin{multline}
    \mathcal{L}_{adv}(G,D) =  \mathbb{E}_{x,y} \big[\ (D(x,y) -1)^2 \big]\ +  \\ \mathbb{E}_{x,y} \big[\ (D(G(x),y)+1))^2 \big]\
\label{eq1}
\end{multline}
In Eq.~\ref{eq1}, we first train the discriminators on the real ECG signals, $x$. After that, we train with the synthesized signal, $G(x)$. We begin by batch-wise training the discriminators,$D$ on the training data. Following that, we train the $G$ while keeping the weights of the discriminators frozen. In a similar fashion, we train $G$ on a batch of training samples while keeping the weights of all the discriminators frozen. 

For classification of different normal and arrhythmia signals, we use categorical cross-entropy as in Eq.~\ref{eq2}.
\begin{equation}
    \mathcal{L}_{class}(D) = -\sum^{k}_{i=0} y_i\log(y'_i)
\label{eq2}
\end{equation}
The generators also incorporate the reconstruction loss as shown in Eq.~\ref{eq3}. We ensure that the synthesized signal contains representative features and bounded by the L2 reconstruction loss. This in turn helps with the generator to output realistic signals with small perturbations and the discriminator to be robust against such adversarial examples, as shown in \cite{meng2017magnet,samangouei2018defense}.

\begin{equation}
    \mathcal{L}_{rec}(G) = \mathbb{E}_{x} \Vert G(x) - x \Vert^2
    \label{eq3}
\end{equation}

By incorporating Eq.~\ref{eq1}, \ref{eq2}  we can formulate our final objective function as given in Eq.~\ref{eq4}.
\begin{multline}
\min \limits_{G,D} \big( \max \limits_{D}  (\mathcal{L}_{adv}(G,D)) + \lambda_{rec}\big[\ \mathcal{L}_{rec}(G_{f},G_{c})\big]\ +  \\
\lambda_{class}\big[\ \mathcal{L}_{class}(D)\big]\ \big)    
\label{eq4}
\end{multline}

Here,  $\lambda_{rec}$, and $\lambda_{class}$ denote different weights, that are multiplied with their respective losses. The loss weighting decides which architecture to prioritize while training.

\section{Experiment}
\iffalse
In the following section, different experimentation and evaluation are provided for our proposed architecture. First, we elaborate on the data preparation in 3.1 and pre-processing scheme in Sec. 3.2. We then define our hyper-parameter settings in Sec. 3.3. Different architectures are compared based on some quantitative(Sec. 3.4) and qualitative (Sec. 3.5) evaluation metrics for real examples and adversarial examples. 
\fi

\begin{figure*}[htp]
    \centering
    \includegraphics[width=0.9\textwidth]{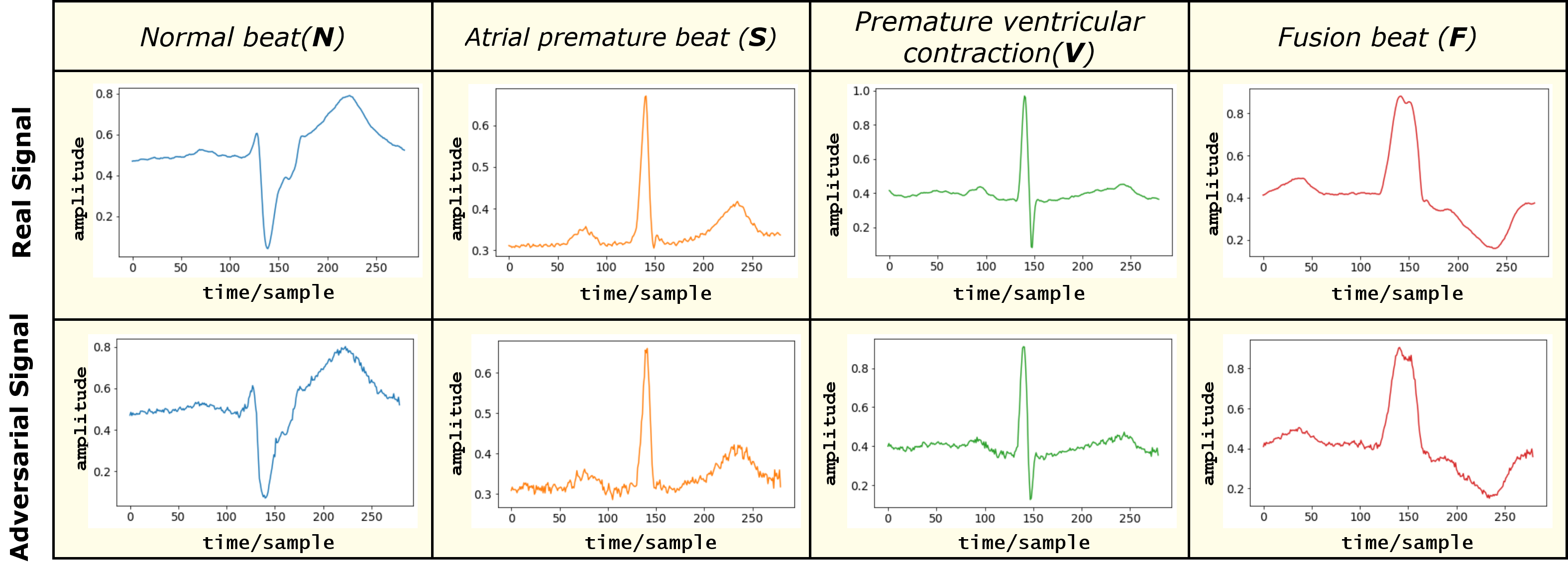}
    \caption{First row contains Real ECG signals for different classes (N,S,V,F). Second row contains adversarial ECG signals generated from the real ones. For each graph, the X-axis signifies sample/time in range of $[0,280]$ and Y-axis signifies amplitude of $[0,1]$}
    \label{fig2}
\end{figure*}
\subsection{Data Set}

PhysioNet MIT-BIH Arrhythmia dataset includes the ECG signals for 48 subjects, including 25 men and 22 women, recorded at the sampling rate of 360Hz. American association of medical instrumentation (AAMI) recommends the database as it contains the five fundamental arrhythmia groups. For our experiment, we divided the dataset into four categories, \textbf N [Normal beat (N), Left and right bundle branch block beats (L, R), Atrial escape beat (e), Nodal (junctional) escape beat (j)], \textbf S [Atrial premature beat (A), Aberrated atrial premature beat (a), Nodal (junctional) premature beat (J), Supraventricular premature beat (S)], \textbf V [Premature ventricular contraction (V), Ventricular escape beat (E)],\textbf F [Fusion of the ventricular and normal beat (F)]. We used lead II between the two leads in ECG signals, collected by placing the electrodes on the chest.

\subsection{Signal Pre-processing and Data Preparation}
We first find the R-Peak for every signal and use sampling rate of 280 centering on R-Peak. We then normalize the amplitude of the signals between $[0,1]$. To test our model's robustness against adversarial examples, we divided the original data into two 1) inter-patient and 2) intra-patient sets, as done in \cite{mousavi2019inter}. In the intra-patient data, we randomly choose the training and the test sets from the same patient records. In contrast, inter-patient data consists of training and testing set with different patient records. Following Chazal et al.\cite{de2004automatic}, we divided the dataset into subsets of DS1 ={101, 101, 106, 108,109, 112, 114, 115, 116, 118, 119, 122, 124, 201, 203, 205, 207, 208, 209, 215, 220, 223,230} and DS2 = {100, 103, 105, 111, 113, 117, 121, 123, 200, 202, 210, 212, 213, 214, 219, 221, 222, 228, 231, 232, 233, 234}. In the Intra-patient benchmarking,  we combine and randomize DS1 and DS2. Then we divide that into 80\% and 20\% sets of train and test data. For the Inter-patient benchmarking, we train on the DS1 dataset and test on the DS2 dataset. From Intra-patient data, we take  N: 87529, S: 5892, V: 2438, F: 2438 samples and divide this into 80\% and 20\% folds. In the case of Inter-patient data, we train on the DS1 and test on DS2. The sample size for train data is N: 44476, S: 3205, V: 683, and test data is N: 43053, S: 2687, V: 1755. We can see that there is a lack of samples for minority classes. To address this, we use Synthetic Minority Over-sampling Technique (SMOTE) to balance the number of samples for training \cite{chawla2002smote}.

\begin{table}[tbp]
\caption{\textbf{Generator's Performance}: Similarity between adversarial examples and Inter-Patient and Intra-Patient test set.}
\begin{adjustbox}{}
\begin{tabular}{|c|c|c|c|c|}
\hline
Dataset                                                  & MSE     & \begin{tabular}[c]{@{}c@{}}Structural\\ Similarity\end{tabular} & \begin{tabular}[c]{@{}c@{}}Cross-corelation\\ Coefficient\end{tabular} & \begin{tabular}[c]{@{}c@{}}Normalized\\  RMSE\end{tabular} \\ \hline
\begin{tabular}[c]{@{}c@{}}Intra \\ Patient\end{tabular} & 0.00241 & 0.9977                                                          & 0.9425                                                                 & 0.0292                                                     \\ \hline
\begin{tabular}[c]{@{}c@{}}Inter\\ Patient\end{tabular}  & 0.00038 & 0.9982                                                          & 0.9986                                                                 & 0.0136                                                     \\ \hline
\end{tabular}
\end{adjustbox}
\label{table1}
\end{table}

\subsection{Hyper-parameters}
For adversarial training, we used least-square-GAN \cite{mao2017least}. We chose $ \lambda_{class} =10$ and $ \lambda_{rec} =1$ (Eq.~\ref{eq4}). For optimizer, we used Adam with a learning rate of $\alpha=0.0002$, $\beta_1=0.5$ and $\beta_2=0.999$. We used Tensorflow 2.0 to train the model in mini-batches with the batch size, $b=64$ in 200 epochs which took around 8 hours to train on NVIDIA RTX2080 GPU. We generated random float values in a range of $[0,1]$ for initializing the noise vector, z. The inference time of our generator and discriminator are 0.0987 and 0.053 milliseconds per instance. 

\begin{table*}[htp]
\centering
\caption{\textbf{Intra-patient ECG classification} : Comparison of architectures trained and tested on \textbf{real} Intra-Patient MIT-BIH dataset and further evaluated on adversarial examples, generated from Intra-Patient test set.}
\begin{adjustbox}{width=1\textwidth}
\begin{tabular}{|c|c|c|c|c|c|c|c|c|c|c|}
\hline
\multirow{2}{*}{Dataset} &\multirow{2}{*}{Method} & \multirow{2}{*}{ACC} &\multicolumn{2}{c|}{N}  & \multicolumn{2}{c|}{S}  & \multicolumn{2}{c|}{V} & \multicolumn{2}{c|}{F} \\\cline{4-11} 
& &  & SEN & SPEC & SEN & SPEC & SEN & SPEC & SEN & SPEC \\ 

\hline\hline
\multirow{8}{*}{\begin{tabular}[c]{@{}c@{}}Real ECG\\ Intra-Patient\end{tabular}}& \textbf{Proposed Method}  & 99.31  & 99.28  & 94.36 & 94.56 & 99.64 & 94.93 & 99.94 & 95.54 & 99.54\\ \cline{2-11} 

&Mousavi et al.\cite{mousavi2019inter} & 99.92 & 100.0 &  98.87 &  96.48 & 100.00 & 99.50 & 99.98 & 98.68 & 99.98\\ \cline{2-11} 
&Shaker et al. \cite{shaker2020generalization}   & 98.00 & 96.29  & 99.68 & 99.30 & 97.71 & 99.74 & 98.55 & 99.80 & 96.95\\ \cline{2-11} 
&Kachuee et al. \cite{kachuee2018ecg}& 93.4 & - & - & - & - & - & - & - & -\\\cline{2-11} 
&Acharya et al. \cite{acharya2017deep}& 97.37 & 91.64 & 96.01 & 89.04 & 98.77 & 94.07 & 98.74 & 95.21 & 98.67 \\\cline{2-11} 
&Ye et al. \cite{ye2010arrhythmia}& 96.50 & 98.7 & - & 72.4 & - & 82.6 & - & 65.6 & - \\\cline{2-11} 
&Yu and Chou \cite{yu2008integration}& 95.4 & 96.9 & - & 73.8 & - & 92.3 & - & 51.0 & - \\\cline{2-11} 
&Song et al \cite{song2005support}& 98.7 & 99.5 & - & 86.4 & - & 95.8 & - & 73.6 & -\\
\hline	\hline
\multirow{5}{*}{\begin{tabular}[c]{@{}c@{}}Adversarial ECG\\ Intra-Patient\end{tabular}}& \textbf{Proposed Method}  & \textbf{99.89} & \textbf{99.96} & 98.63  & \textbf{95.56} & \textbf{99.98} & \textbf{99.02} & \textbf{99.99} & \textbf{100.0} & \textbf{99.94} \\ \cline{2-11} 
&Mousavi et al. \cite{mousavi2019inter}   & 86.97 &  72.75 & 99.31 & 92.81 & 93.24 & 81.26 & 98.90 & 100.0 & 81.48\\ \cline{2-11} 
&Shaker et al. \cite{shaker2020generalization}   & 80.33 & 59.09  & 90.71 & 88.58 & 91.50 & 68.02 & 98.93 & 99.12 & 68.90\\ \cline{2-11} 
&Acharya et al. \cite{acharya2017deep}  & 69.58 & 35.50  & 98.69 & 71.45 & 84.32 & 74.33 & 97.29 & 99.36 & 56.78\\\cline{2-11} 
&Kachuee et al. \cite{kachuee2018ecg}  & 75.91 & 49.47  & \textbf{99.65} &  79.49 &  95.22 & 70.33 & 96.53 & 100.0 & 59.45\\ 
\hline
\end{tabular}
\end{adjustbox}
\label{table2}
\end{table*}

\begin{table*}[htp]
\centering
\caption{\textbf{
Inter-patient ECG classification} : Comparison of architectures trained and tested on \textbf{real} Inter-Patient MIT-BIH dataset and further evaluated on adversarial examples, generated from Inter-Patient test set.}

\begin{adjustbox}{width=1\textwidth}
\begin{tabular}{|c|c|c|c|c|c|c|c|c|c|c|}
\hline
\multirow{2}{*}{Dataset}&\multirow{2}{*}{Method} & \multirow{2}{*}{ACC} &\multicolumn{2}{c|}{N}  & \multicolumn{2}{c|}{S}  & \multicolumn{2}{c|}{V} & \multicolumn{2}{c|}{F} \\\cline{4-11} 
& &  & SEN & SPEC & SEN & SPEC & SEN & SPEC & SEN & SPEC \\ 
\hline
\hline
\multirow{7}{*}{\begin{tabular}[c]{@{}c@{}}Real ECG\\ Inter-Patient\end{tabular}}& \textbf{Proposed Method} &  97.72 & 95.87 & 96.00 & 88.19  & 99.60 & 93.75 &  99.64 &  90.90 & 96.53\\ 

\cline{2-11}
& Mousavi et al.\cite{mousavi2019inter} & 99.53 & 99.68 &  96.05 &  88.94 &  99.72 & 99.94 &  99.97 & - & -\\ 

\cline{2-11}
& Garcia et al. \cite{garcia2017inter}& 92.4 & 94.0 & 82.6 & 62.0 & 97.9 & 87.3 & 95.9 & - & -\\

\cline{2-11}
& Lin and Yang. \cite{lin2014heartbeat}& 96.50 & 91.0 & - & 81.0 & - & 86.0 & - & - & - \\
 
\cline{2-11}
& Ye et al. \cite{ye2010arrhythmia}& 75.2 & 80.2 & - & 3.2 & - & 50.2 & - & - & -\\

\cline{2-11}
& Yu and Chou \cite{yu2008integration}& 75.2 & 78.3 & - & 1.8 & - & 83.9 & - & 0.3 & - \\

\cline{2-11}
& Song et al \cite{song2005support}& 76.3 & 78.0 & - & 27.0 & - & 80.8 & - & 73.6 & - \\
\hline	
\hline

\multirow{5}{*}{\begin{tabular}[c]{@{}c@{}}Adversarial ECG\\ Inter-Patient\end{tabular}}& \textbf{Proposed Method}  & \textbf{98.39} & \textbf{97.52}  & 94.06 & 93.54 & \textbf{99.75} & \textbf{93.18} & \textbf{99.85} & 93.50 & \textbf{97.71}\\ 
\cline{2-11}
& Mousavi et al. \cite{mousavi2019inter}   & 75.90 & 48.65 & 99.02 & 90.37  &  90.90 & 73.10 & 99.80 & 100.0 & 60.57\\
\cline{2-11}
& Shaker et al. \cite{shaker2020generalization}   & 81.93 & 61.31  & 98.92 & 90.37 & 96.28 & 84.65 & 99.56 & 97.40 & 67.71\\ 
\cline{2-11}
& Acharya et al. \cite{acharya2017deep}  & 79.63 & 57.26  & \textbf{99.78} & \textbf{95.03} & 73.23 & 64.58 & 98.02 & \textbf{98.70} & 86.92\\ \cline{2-11}
& Kachuee et al. \cite{kachuee2018ecg}  & 66.52 & 31.79  & 99.46 &  22.36 & 94.39 & 50.75 & 94.63 & 97.40 & 43.13\\ 
\hline	
\end{tabular}
\end{adjustbox}
\label{table3}
\end{table*}

\subsection{Qualitative Evaluation}
% ***********Intra Patient************
For finding the structural similarity with the original ECG, we benchmark synthesized adversarial signals using four different metrics, i) Mean Squared Error (MSE), ii) Structural Similarity Index (SSIM), iii) Cross-correlation coefficient, and iv) Normalized Mean Squared Error (NRMSE). We use the same Intra and Inter Patient test set for synthesizing adversarial examples. Table.~\ref{table1} shows that SSIM for both Inter and Intra Patient has a 99.8\% score, which means the adversarial examples are structurally similar to the original signal. As for cross-correlation, MSE, and NRMSE, our model generates better Inter-patient adversarial examples over Intra-patient ones. Thus the score for Inter Patient test set is better than the Intra Patient dataset. It is important to note that we want to achieve lower MSE and RMSE. Similarly, we want to score higher for SSIM and Cross-correlation coefficient. For visual comparison, we provide real and synthesized ECG signals for each class in Fig.~\ref{fig2}. The columns contain Normal (N), Atrial Premature (A), Premature Ventricular (V), and Fusion (F) beats. In contrast, the row contains Real and Adversarial Signals successively. It is apparent from the figure that the synthesized signals are quite vivid and realistic visually.

 \iffalse
\begin{table*}[htp]
\centering
\caption{\textbf{Adversarial Intra-Patient ECG classification}: Comparison of architectures tested on adversarial examples, generated from Intra-Patient test set.}
\begin{adjustbox}{width=0.8\textwidth}
\begin{tabular}{|c|c|c|c|c|c|c|c|c|c|}
\hline
\multirow{2}{*}{Method} & \multirow{2}{*}{ACC} &\multicolumn{2}{c|}{N}  & \multicolumn{2}{c|}{S}  & \multicolumn{2}{c|}{V} & \multicolumn{2}{c|}{F} \\\cline{3-10} 
&  & SEN & SPEC & SEN & SPEC & SEN & SPEC & SEN & SPEC \\ 

\hline
Proposed Method  & \textbf{99.89} & \textbf{99.96} & \textbf{98.63}  & \textbf{95.56} & \textbf{99.98} & \textbf{99.02} & \textbf{99.99} & \textbf{100.0} & \textbf{99.94} \\ 
\hline	
Mousavi et al. \cite{mousavi2019inter}   & 86.97 &  72.75 & 99.31 & 92.81 & 93.24 & 81.26 & 98.90 & 100.0 & 81.48\\ 
\hline
Shaker et al. \cite{shaker2020generalization}   & 80.33 & 59.09  & 90.71 & 88.58 & 91.50 & 68.02 & 98.93 & 99.12 & 68.90\\ 
\hline
Acharya et al. \cite{acharya2017deep}  & 69.58 & 35.50  & 98.69 & 71.45 & 84.32 & 74.33 & 97.29 & 99.36 & 56.78\\ \hline
Kachuee et al. \cite{kachuee2018ecg}  & 75.91 & 49.47  & 99.65 &  79.49 &  95.22 & 70.33 & 96.53 & 100.0 & 59.45\\ 
\hline
\end{tabular}
\end{adjustbox}
\label{table3}
\end{table*}
\fi

\begin{table*}[htp]
\caption{\textbf{Adversarial Example Detection}: Evaluating the discriminator's performance for detecting adversarial example on Intra and Inter Patient Data}
\centering
\begin{adjustbox}{width=0.9\linewidth}

\begin{tabular}{|c|c|c|c|c|c|c|c|c|}
\hline
\multirow{2}{*}{Dataset} & \multirow{2}{*}{Accuracy} &\multicolumn{2}{c|}{Sensitivity}  & \multicolumn{2}{c|}{Precision} & \multicolumn{2}{c|}{ F1 Score} & \multirow{2}{*}{AUC}\\\cline{3-8} 
&  & Real & Adversarial & Real & Adversarial & Real & Adversarial &\\ 

\hline
Intra-Patient &  87.83  & 82.93 & 92.74 & 91.95   &   84.45  & 87.21 & 88.40 & 87.83
\\ \hline
Inter-Patient &   77.82    & 64.48  & 91.17   &  74.41  & 80.44 &  87.96 & 71.96 & 77.82 \\
\hline
\end{tabular}
\end{adjustbox}
\label{table6}
\end{table*}
\begin{figure*}[htbp]
    \centering
    \includegraphics[height=13cm,width=0.9\textwidth]{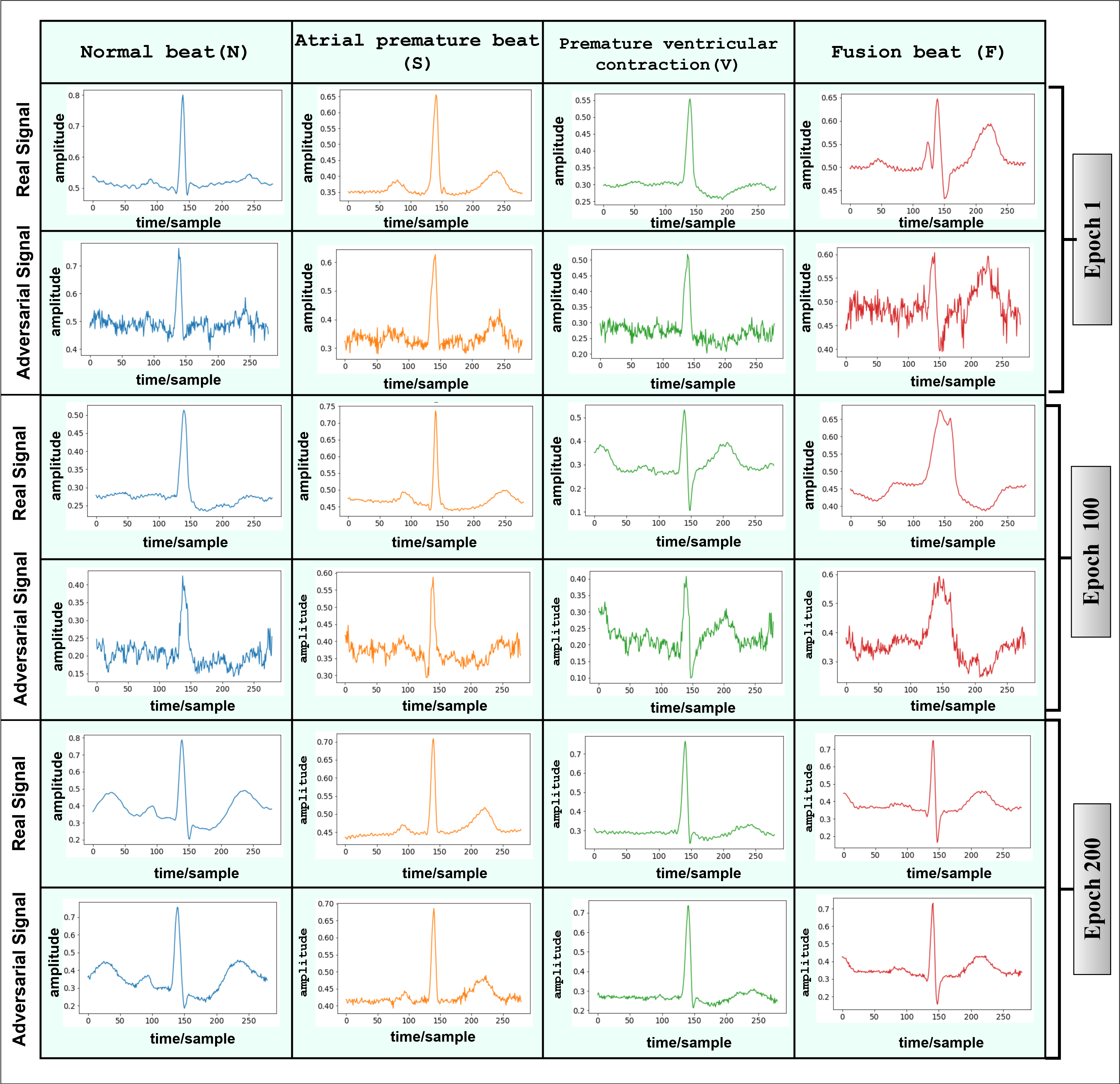}
    \caption{Each pair of row contains real and adversarial signal for Epoch 1,100 and 200 successively.The column consists of different ECG signals such as Normal (N), Atrial Premature (A), Premature Ventricular (V), and Fusion (F) beats and for each graph, the X-axis signifies \textbf{sample/time} in range of $[0,280]$ and Y-axis signifies \textbf{amplitude} of $[0,1]$.We can see that our ECG-adv-GAN synthesizes realistic looking adversarial ECGs over time}
    \label{fig3}
\end{figure*}

\subsection{Quantitative Evaluation}
For normal and arrhythmia beat classification tasks, we compare our model with other state-of-the-art architectures on Intra-Patient and Inter-Patient test sets as given in Table.~\ref{table2} and Table.~\ref{table3}. We further experiment on adversarial examples synthesized using our Generator from the above two tests set, which we also provide in Table.~\ref{table2} and Table.~\ref{table3}. For metrics, we use Accuracy (ACC), Sensitivity(SEN), and Specificity(SPEC). We can see for the first experiment, except for the architecture given in \cite{mousavi2019inter}, our model achieves the best score compared to other deep learning and machine learning derived architectures. The architecture in \cite{mousavi2019inter} uses Seq2Seq, consisting of Bidirectional LSTM and RNN layers. On the other hand,  \cite{shaker2020generalization,acharya2017deep,kachuee2018ecg} uses 1D Convolution based architecture. Out of these models, Shaker et al. \cite{shaker2020generalization} adopt DC-GAN, a generative network \cite{radford2015unsupervised} for adversarial signal generation. However, their classification architecture is trained separately, and they provide results only on real ECG signals. 

As for the second experiment, we evaluate our proposed ECG-Adv-GAN and other architectures on adversarial examples generated from the Intra-patient and Inter-patient data-set. The authors of these architectures failed to provide any pre-trained weight that we can use to evaluate. So for a fair comparison, we train and test them on the same set. This evaluation's main idea is to test the model's effectiveness and robustness on out-of-distribution and real-world data, which the adversarial example imitates. As shown in Table.~\ref{table2}, our model outperforms other the proposed methods for the Intra-Patient data-set. In Table.~\ref{table3} for the Inter-Patient data-set, our model achieves the best overall accuracy and sensitivity across all the normal and abnormal beat categories. Only the architecture by Acharya et al. \cite{acharya2017deep} scores better specificity for Normal and Fusion beats. We can also see that Mousavi et al. \cite{mousavi2019inter} performs worse for out-of-distribution adversarial data-set, which confirms signs of over-fitting on the real ECG signals.

\iffalse
\begin{table*}[htp]
\centering
\caption{\textbf{Adversarial Inter-Patient ECG classification}: Comparison of architectures tested on adversarial examples, generated from Inter-Patient test set.}
\begin{adjustbox}{width=1\textwidth}
\begin{tabular}{|c|c|c|c|c|c|c|c|c|c|}
\hline
\multirow{2}{*}{Method} & \multirow{2}{*}{ACC} &\multicolumn{2}{c|}{N}  & \multicolumn{2}{c|}{S}  & \multicolumn{2}{c|}{V} & \multicolumn{2}{c}{F} \\\cline{3-10} 
&  & SEN & SPEC & SEN & SPEC & SEN & SPEC & SEN & SPEC \\ 

\hline
Proposed Method  & \textbf{98.39} & \textbf{97.52}  & 94.06 & \textbf{93.54} & \textbf{99.75} & \textbf{93.18} & \textbf{99.85} & 93.50 & \textbf{97.71}\\ 
\hline 
Mousavi et al. \cite{shaker2020generalization}   & 75.90 & 48.65 & 99.02 & 90.37  &  90.90 & 73.10 & 99.80 & 100.0 & 60.57\\
\hline
Shaker et al. \cite{shaker2020generalization}   & 81.93 & 61.31  & 98.92 & 90.37 & 96.28 & 84.65 & 99.56 & 97.40 & 67.71\\ 
\hline
Acharya et al. \cite{acharya2017deep}  & 79.63 & 57.26  & \textbf{99.78} & 95.03 & 73.23 & 64.58 & 98.02 & \textbf{98.70} & 86.92\\ \hline
Kachuee et al. \cite{kachuee2018ecg}  & 66.52 & 31.79  & 99.46 &  22.36 & 94.39 & 50.75 & 94.63 & 97.40 & 43.13\\ 
\hline	
\end{tabular}
\end{adjustbox}
\label{table5}
\end{table*}
\fi

For the third experimentation,  we evaluate our discriminator and check its accuracy to detect adversarial examples as given in Table.~\ref{table6}. We use a test set with 50\% adversarial and 50\%  real data to carry out this benchmark. By combining the real and the adversarial examples synthesized by our generator, we can have this 50/50 test split. So for the Intra-Patient and  Inter-Patient set of 19327 and 19152 signals, we end up having 38654 and 38304 samples successively. We use Accuracy, Sensitivity, Precision, F1-score, and AUC as standard metrics for measuring our model's performance. As shown in Table.~\ref{table6}, both our model scores high accuracy, precision, sensitivity, and AUC  for the Intra-Patient and Inter-Patient data-set. However, the model trained on Intra-Patient data performs better for detecting both and real adversarial examples with more than 80\% F1-score and Precision. Another important insight is that the model detects adversarial examples far better than real ECG signals. Consequently, it confirms the hypothesis that Generative Networks are far superior for detecting adversarial examples.

% ***********Benchmarking************

\subsection{Stability of Signal Generation}

To evaluate the stability of GAN during training time, we illustrate distinct signals synthesized by the generator network in Figure~\ref{fig3}. In the early stage of training, the ECG signal's complexity sometimes prevents the generator from learning a select number of features to fool the discriminator network, as seen in the first row of the image (Epoch 1). Over time, we can see that the outputs synthesized by the generator become more visually realistic. For example, on Epoch-200, the peak of the signal for every class (N, S, V, F) is not differentiable from the original ECG-Signal.  To validate, we monitored the GAN's stability by visually evaluating the synthesized samples for each epoch. The signals in Figure \ref{fig3} produced by the ECG-Adv-GAN were tested with our Discriminator and other state-of-the-art classifiers to determine if the models can accurately classify adversarial samples. From Tables \ref{table2} and \ref{table3}, we can see that our Discriminator outperforms other architecture in classifying adversarial examples.

\section{Conclusion}
In this paper, we introduced ECG-Adv-GAN, a novel conditional Generative Adversarial Network for simultaneous synthesis of adversarial examples and detecting arrhythmia.  Our architecture outperforms previous techniques by adopting a dual learning task of synthesizing adversarial ECG signals while predicting the signal category. The model is best suited for real-time ECG monitoring, where it can perform robustly and effectively. One future direction to this work is to incorporate and defend against adversarial attacks in ECG signals.

\bibliographystyle{IEEEtran}
\bibliography{reference}
\end{document}